%% file: root.tex
 \documentclass[letterpaper, 10 pt, journal, twoside]{IEEEtran}

\IEEEoverridecommandlockouts                              

\usepackage{graphics} 
\usepackage{graphicx}
\usepackage{epsfig} 
\usepackage{amsmath,nccmath} 
\usepackage{amssymb}  
\usepackage{cite}
\usepackage{multirow}
\usepackage{booktabs}
\usepackage{multicol}
\usepackage[table]{xcolor}
\usepackage{color,soul}
\usepackage{acronym}
\usepackage{mathrsfs}
\usepackage[font=footnotesize]{caption}
\usepackage{nicefrac}

\input{notation_header}

\acrodef{ESGVI}{Exactly Sparse Gaussian Variational Inference}
\acrodef{ELBO}{Evidence Lower Bound}
\acrodef{EM}{Expectation-Maximization}
\acrodef{MAP}{Maximum A Posteriori}
\acrodef{WNOA}{white-noise-on-acceleration}
\acrodef{2D}{2-dimensional}
\acrodef{3D}{3-dimensional}
\acrodef{RBF}{Radial Basis Functions}
\acrodef{VAE}{Variational Autoencoder}
\acrodef{SLAM}{Simultaneous Localization and Mapping}
\acrodef{ICP}{Iterative Closest Point}
\acrodef{LOAM}{Lidar Odometry and Mapping}
\acrodef{SuMa}{Surfel-based Mapping}

\title{Unsupervised Learning of Lidar Features for Use in a Probabilistic Trajectory Estimator}

\author{David J. Yoon$^{1}$, Haowei Zhang$^{1}$, Mona Gridseth$^{1}$, Hugues Thomas$^{1}$, Timothy D. Barfoot$^{1}$%
  \thanks{Manuscript received: October 14, 2020; Revised January 7, 2021; Accepted January 28, 2021.}
  \thanks{This paper was recommended for publication by Editor Sven Behnke upon evaluation of the Associate Editor and Reviewers' comments. This work was supported by Applanix Corporation and the Natural Sciences and Engineering Research Council of Canada (NSERC).} 
  \thanks{$^{1}$All authors are with the University of Toronto Institute for Aerospace Studies (UTIAS), Canada.
    {\tt\scriptsize \{david.yoon, gareth.zhang, mona.gridseth, hugues.thomas\}@robotics.utias.utoronto.ca, tim.barfoot@utoronto.ca}}%
  \thanks{Digital Object Identifier (DOI): see top of this page.}
}

\begin{document}

\maketitle

\markboth{IEEE Robotics and Automation Letters. Preprint Version. Accepted January, 2021}
{Yoon \MakeLowercase{\textit{et al.}}: Unsupervised Learning of Lidar Features for Use in a Probabilistic Trajectory Estimator} 

\begin{abstract}
We present unsupervised parameter learning in a Gaussian variational inference setting that combines classic trajectory estimation for mobile robots with deep learning for rich sensor data, all under a single learning objective. The framework is an extension of an existing system identification method that optimizes for the observed data likelihood, which we improve with modern advances in batch trajectory estimation and deep learning. Though the framework is general to any form of parameter learning and sensor modality, we demonstrate application to feature and uncertainty learning with a deep network for 3D lidar odometry. Our framework learns from only the on-board lidar data, and does not require any form of groundtruth supervision. We demonstrate that our lidar odometry performs better than existing methods that learn the full estimator with a deep network, and comparable to state-of-the-art ICP-based methods on the KITTI odometry dataset. We additionally show results on lidar data from the Oxford RobotCar dataset.
\end{abstract}

\begin{IEEEkeywords}
  Localization, Deep Learning Methods, Field Robots
\end{IEEEkeywords}


\section{INTRODUCTION}
\IEEEPARstart{P}{robabilistic} state estimation is a mature component of autonomous navigation. Large estimation problems with rich data (e.g., lidar, camera) can be solved efficiently by exploiting the sparse structure inherent to the probabilistic formulation. However, the implementation of estimators can vary depending on the platform, sensor, and application environment. Rather than requiring expert engineers to adapt existing estimators for new deployments, our vision is to develop a learning framework that can learn model parameters specific to the deployment, solely from the sensor data.

Recently, Barfoot et al. \cite{barfoot_ijrr20} presented \ac{ESGVI}, a nonlinear batch state estimation framework that starts from a variational objective, and provides a family of scalable estimators by exploiting the factorization of the joint likelihood between the observed measurements (data) and state. Wong et al. \cite{wong_ral20b} demonstrated the extension of \ac{ESGVI} to parameter learning, and successfully learned robot noise models that are robust to noisy measurements and outliers, solely from the observed measurements (i.e., without groundtruth).

In this paper, we demonstrate parameter learning for a deep neural network with \ac{ESGVI} for the first time, resulting in a novel hybrid of deep learning and probabilistic state estimation. We apply our method to feature and uncertainty learning for lidar odometry with a network architecture built using KPConv \cite{thomas2019KPConv}, a pointcloud convolution operator. 

We do not train with groundtruth supervision; all trainable network parameters are learned solely from the on-board lidar data. Experimental results on Velodyne HDL-64 lidar data of the KITTI odometry dataset \cite{Geiger2012} show that our method performs better than those that learn the estimator with a deep network, and comparable to the current state of the art, i.e., \ac{ICP}-based methods. We additionally show results on Velodyne HDL-32 data from the Oxford RobotCar dataset \cite{Maddern2017, barnes2020oxford}, where we demonstrate the simplicity in retraining the network for different deployment regions.

We review related work in Section \ref{sec:related_work}, and provide an overview of \ac{ESGVI} parameter learning in Section \ref{sec:esgvi}. Section \ref{sec:method} is the methodology of our lidar odometry, and experimental results are presented in Section \ref{sec:experiments}. Finally, we discuss concluding remarks and future work in Section \ref{sec:conclusion}.

\begin{figure}[!t]
  \centering
  \includegraphics[width=0.49\textwidth]{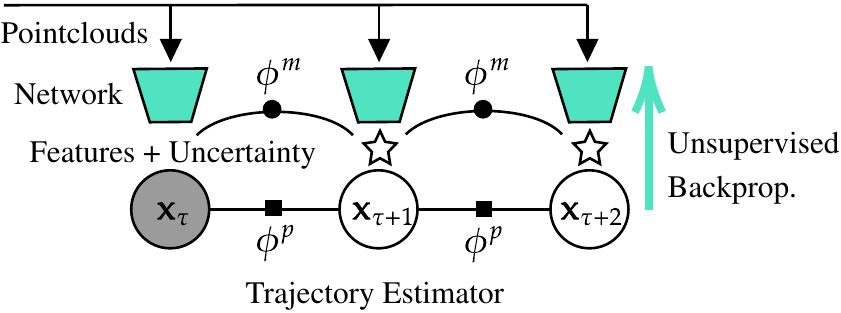}
  \caption{An example factor graph diagram of a lidar odometry problem. We optimize the trajectory over a sliding window of $w$ time frames (e.g., $w=3$ above), where $\mbf{x}_k$ is our state at time $t_k$. The first frame is locked (grey) and is the reference frame, we do not optimize it. Each frame receives a pointcloud from the lidar sensor. A deep network takes each pointcloud as input and outputs features with uncertainty (stars) that can be associated to other frames and composed into measurement factors, $\phi^m$ (circles). Motion prior factors, $\phi^p$ (squares), are applied between every frame. We do not require supervision, and only learn from the on-board lidar data.}
  \label{fig:factor_graph}
  \vspace*{-0.2in}
\end{figure}

\section{Related Work}
\label{sec:related_work}

Parameter learning with \ac{ESGVI} originates from a linear system identification method, where Ghahramani and Hinton \cite{ghahramani96} optimize the likelihood of the observed measurements (data) by introducing a latent trajectory (state) and applying \ac{EM}. In the E-step, model parameters are held fixed and the trajectory is optimized with Kalman smoothing. In the M-step, the trajectory is held fixed and the model parameters are optimized. Critically, this method is able to learn entire linear models from just the observed data, with no prior knowledge. Ghahramani and Roweis \cite{ghahramani99} extend this concept to simple nonlinear models approximated with Gaussian \ac{RBF}.

Barfoot et al. \cite{barfoot_ijrr20} recently presented a new nonlinear estimation framework that exploits the sparsity in not just smoothing problems (i.e., linear chain), but any factorization of the joint likelihood between data and state (i.e., cycles in the factor graph). Wong et al. \cite{wong_ral20b} apply \ac{ESGVI} estimation to parameter learning with \ac{EM}, thus advancing the \ac{EM} learning framework to modern advances in nonlinear batch estimation. They were able to learn robot noise models robust to outliers, including robustness to false loop closures for pose-graph optimization. Our work further extends this framework to deep learning in order to directly handle rich sensor data such as lidar.

A similar concept based on optimizing the data likelihood by introducing a latent state is applied in the \ac{VAE} \cite{kingma2013auto} framework. With \ac{VAE}s, inference of the latent state is approximated as a deep network (i.e., data input maps to state output), so the bound on the data likelihood can be optimized without \ac{EM}. This approximation is restrictive for time-series application, such as trajectory estimation. \ac{EM} parameter learning with \ac{ESGVI} gives us all the benefits of classic probabilistic estimation, such as information propagation over the entire latent trajectory, and the flexibility of multiple sensors as additional factors.

Alternatively, Bloesch et al. \cite{bloesch2018codeslam} use a \ac{VAE} in probabilistic trajectory estimation without directly inferencing the state with a deep network. They train a \ac{VAE} to learn an efficient (lower-dimensional) latent space for geometry, and optimize over this domain jointly with pose variables at test time for monocular vision estimation problems. Czarnowski et al. \cite{czarnowski2020deepfactors} extend this work by using the same depth representation in a full dense \ac{SLAM} system. Unlike our approach, their network is trained independently from the trajectory estimator, and a training dataset with groundtruth depth images is required.

Evidently, deep learning with core components of probabilistic estimation is an increasingly popular avenue of research. Tang and Tan \cite{tang2018ba} maintain the differentiability of the Levenberg-Marquardt optimizer by iterating for a fixed number of steps and proposing a network to predict the damping factor. Similarly, Stumberg et al. \cite{von2020gn} backpropagate through the Gauss-Newton update step from a random initial condition. Jatavallabhula et al. \cite{jatavallabhula2019} go even further by proposing differentiable alternatives to all modules in full \ac{SLAM} systems as computational graphs. In contrast, our approach does not rely on making the estimator differentiable and so facilitates using any probabilistic estimation method. 

Most similar to our method is the work of DeTone et al. \cite{detone2018b}, where they alternate between training a deep network frontend that outputs visual features from images, and a bundle adjustment backend that optimizes the feature observations as landmarks. The optimized landmarks become the training signal for learning the frontend network. Our approach is derived from a probabilistic objective, and as consequence, our learning objective is different and accounts for uncertainty in the posterior estimates.

While our learning framework is sensor agnostic, our choice of application to demonstrate our work is lidar odometry. Due to the vast amount of literature related to lidar estimation, we restrict our review to the most relevant to our work in the interest of space.

\begin{figure*}
  \centering
  \includegraphics[width=0.9\textwidth]{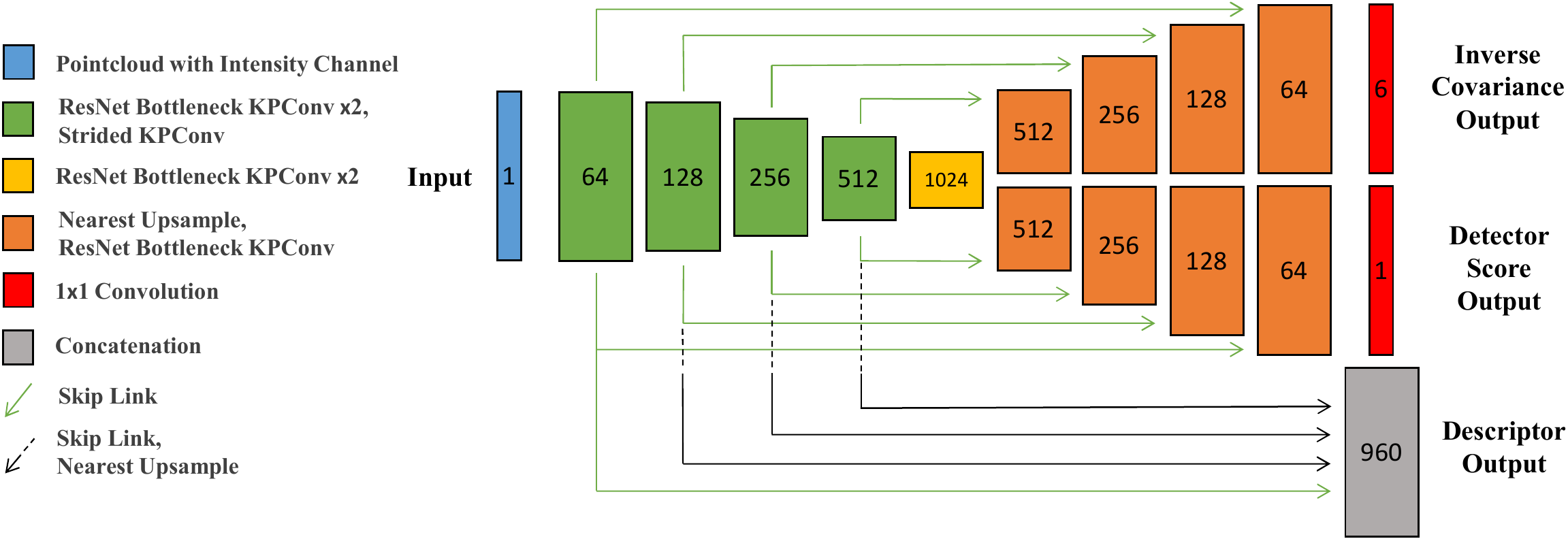}
  \caption{Our network architecture is based on the work of Barnes and Posner \cite{barnes2020}, which we adapt for pointclouds using the KPConv \cite{thomas2019KPConv} pointcloud convolution operator. Input to the network is a lidar pointcloud with an intensity channel. Descriptor vectors for each point are composed from the output of the first four encoder layers. The 6 channel output of the top decoder are composed into inverse measurement covariances (see (\ref{eq:LDL}) in body text). The single channel output of the remaining decoder are detector scores used to compute keypoints (see (\ref{eq:keypoint}) in body text). Refer to the KPConv \cite{thomas2019KPConv} publication for implementation details of the various operations.}
  \label{fig:network}
  \vspace*{-0.2in}
\end{figure*}

The current state of the art for non-learned lidar estimation are those based on \ac{ICP}. Zhang and Singh \cite{zhang2017low} present \ac{LOAM}, which has been the top contender for lidar-only odometry in the KITTI odometry benchmark \cite{Geiger2012} since its inception. Behley and Stachniss \cite{behley2018rss} present \ac{SuMa}, which is notably the method used as the trajectory groundtruth in SemanticKITTI \cite{behley2019iccv}, the KITTI odometry sequences with semantic labels.

At the opposite end, we have fully learned lidar odometry methods that infer relative poses with a deep network. Li et al. \cite{li2019} present LO-Net, a network that takes two pointclouds as input and outputs the relative pose. Their method demonstrates competent odometry performance comparable to \ac{ICP}-based ones, but requires training with supervision from groundtruth trajectory. Cho et al. \cite{cho2020unsupervised} present DeepLO, a network that similarly outputs a relative pose change from two input pointclouds and is trained unsupervised. However, their unsupervised approach comes at a cost, as the odometry performance they present falls short compared to LO-Net and existing \ac{ICP}-based methods.

The learned estimator in LO-Net is impressive, but Li et al. \cite{li2019} demonstrate better odometry in the same publication with \ac{ICP} enhanced with point measurement masks that were trained alongside LO-Net. In similar fashion, Chen et al. \cite{chen2019iros} improve on \ac{SuMa} (\ac{SuMa}++) by incorporating a pretrained semantic classification network. These outcomes suggest that a hybrid of non-learned estimators with learned components can be beneficial. Our work is motivated by the idea that in the spectrum between non-learned and fully learned estimators, there is an optimal balance that can benefit from the advantages of both extremes. The learning framework and odometry solution we present is our attempt at meeting this balance. 

Compared to fully learning the estimator, such as in DeepLO \cite{cho2020unsupervised} and LO-Net \cite{li2019}, our odometry solution achieves better performance while being trained unsupervised. Compared to \ac{SuMa}++ \cite{chen2019iros} and LO-Net with \ac{ICP}, our approach learns more and does not rely on nearest-neighbours for data association. Having more learnable components has the advantage of automation, i.e., being able to tune models from data for different deployments, rather than requiring an expert engineer. Another advantage of not relying on \ac{ICP} is that our method inherently has a good estimate of trajectory uncertainty. Uncertainty estimation for \ac{ICP} is on its own a challenging research problem \cite{landry2019cello, brossard2020new}.

\section{Exactly Sparse Gaussian Variational Inference} \label{sec:esgvi}
In this section, we summarize parameter learning in the \ac{ESGVI} framework as presented by Barfoot et al. \cite{barfoot_ijrr20}. We begin with the maximum-likelihood problem for the given data, $\mbf{z}$, which is expressed as
\begin{equation} \label{eq:data_likelihood}
  \mbs{\theta}^\star = \arg \max_{\mbs{\theta}} p(\mbf{z} | \mbs{\theta}),  
\end{equation}
where $\mbs{\theta}$ represents the parameters of our system that we wish to learn (e.g., parameters of a neural network).

We define the loss to minimize as the negative log-likelihood of the data, $\mathscr{L} = -\ln{p(\mbf{z} | \mbs{\theta})}$, and introduce the latent trajectory, $\mbf{x}$. Applying the usual \ac{EM} decomposition,
\begin{equation}
  \mathscr{L} = \begin{medsize}
  \underbrace{\int^{\mbs{\infty}}_{-\mbs{\infty}} q(\mbf{x}) \ln \left( \frac{p(\mbf{x} | \mbf{z}, \mbs{\theta})}{q(\mbf{x})} \right) d\mbf{x}}_{\mbox{$\leq$ 0}}
  - \underbrace{\int^{\mbs{\infty}}_{-\mbs{\infty}} q(\mbf{x}) \ln \left( \frac{p(\mbf{x}, \mbf{z}  | \mbs{\theta})}{q(\mbf{x})} \right) d\mbf{x}}_{\mbox{upper bound}}, \label{eq:em_decomp} 
  \end{medsize}
\end{equation}
where we define our approximate posterior trajectory as a multivariate Gaussian distribution, $q(\mbf{x}) = \mathcal{N}(\mbs{\mu}, \mbs{\Sigma})$.

While we cannot optimize the first term as it requires computing the true posterior, $p(\mbf{x} | \mbf{z}, \mbs{\theta})$, we can optimize the second (upper bound) term, the so-called (negative) \ac{ELBO}. 

Using the expression for the entropy, $-\int q(\mbf{x}) \ln q(\mbf{x}) d\mbf{x}$, of a Gaussian and dropping constants, the upper bound term can be written as,
\begin{equation}
\label{eq:functional}
V(q|\mbs{\theta}) = \mathbb{E}_q[ \phi(\mbf{x}, \mbf{z}|\mbs{\theta})] + \frac{1}{2} \ln \left( |\mbs{\Sigma}^{-1}| \right),
\end{equation}
where we define $\phi(\mbf{x}, \mbf{z}|\mbs{\theta}) = - \ln p(\mbf{x},\mbf{z}|\mbs{\theta})$, $\mathbb{E}[\cdot]$ is the expectation operator, and $|\cdot|$ is the matrix determinant. As defined by Barfoot et al. \cite{barfoot_ijrr20}, (\ref{eq:functional}) is the \ac{ESGVI} loss functional.

We now apply \ac{EM}\footnote{We work with the negative log-likelihood, therefore we are technically applying Expectation Minimization. However, the acronym stays the same.} and proceed iteratively in two steps: the E-step and the M-step. In the E-step, we hold the parameters, $\mbs{\theta}$, fixed and optimize the posterior estimate $q(\mbf{x})$. In the M-step, we hold the posterior estimate, $q(\mbf{x})$, fixed, and optimize for the parameters, $\mbs{\theta}$. This iterative algorithm gradually optimizes the data likelihood, $\mathscr{L}$, in (\ref{eq:em_decomp}).

Barfoot et al. \cite{barfoot_ijrr20} explain in detail how the E-step can be evaluated efficiently by taking advantage of the factorization of $\phi(\mbf{x}, \mbf{z}|\mbs{\theta})$, the joint likelihood of the state and data. When the expectation over the posterior, $q(\mbf{x})$, is approximated at only the mean of the Gaussian, the E-step is the familiar \ac{MAP} state estimator \cite{barfoot_ijrr20}.

\section{Unsupervised Deep Learning for Lidar Odometry} \label{sec:method}

\subsection{Problem Definition} \label{subsec:problem}
We define our state at time $t_k$ as $\mbf{x}_k = \{\mbf{T}_{k,0}, \mbs{\varpi}_k\}$, where the pose $\mbf{T}_{k,0} \in SE(3)$ is a transformation matrix between frames at $t_k$ and $t_0$, and $\mbs{\varpi}_k \in \mathbb{R}^6$ is the body-centric velocity. We assume we receive a new pointcloud frame from the lidar sensor at each new time $t_k$.

\begin{figure*}
  \centering
  \includegraphics[width=0.96\textwidth]{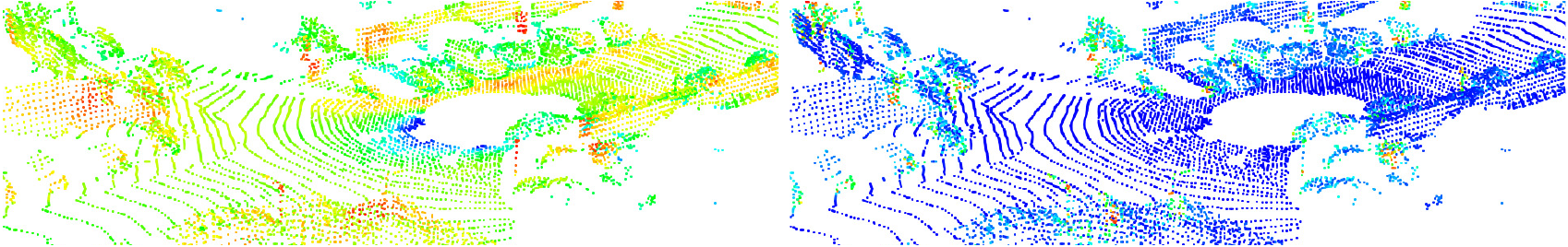}
  \caption{Network outputs coloured in the following order: blue (low value), cyan, green, yellow, and red (high value). (Left) Detector score visualization, highlighting structure such as wall corners and vertical posts. The nearby ground is favoured over vehicles, possibly due to their dynamic nature. (Right) Visualization of sphericity (see Section \ref{subsec:outlier}) computed with the learned measurement covariance. Planar surfaces have low values, the expected result.
  }
  \label{fig:analysis}
  \vspace*{-0.2in}
\end{figure*}

Our odometry implementation is an optimization over a window of $w$ lidar frames, $t_\tau, \dots, t_{\tau+w-1}$. The first pose of the window at $t_\tau$, $\mbf{T}_{\tau,0}$, is locked (not optimized) and treated as the reference frame for keypoint matching. The factorization of our joint likelihood of the state and data is
\begin{equation}
  \phi(\mbf{x}, \mbf{z}|\mbs{\theta}) = \sum_{k=\tau+1}^{\tau+w-1} \left( \phi^p(\mbf{x}_{k-1}, \mbf{x}_k) + \sum_{\ell=1}^{L_k} \phi^m(\mbf{z}_k^\ell | \mbf{x}_\tau, \mbf{x}_k,  \mbs{\theta}) \right),
\end{equation}
where $\mbf{z}_k^\ell$ is the $\ell$th keypoint measurement in lidar frame $k$, which has a total of $L_k$ keypoints. Figure~\ref{fig:factor_graph} shows an example factor graph illustration.

Referring to Figure~\ref{fig:factor_graph}, the square factors, $\phi^p$, are motion prior factors. We apply a white-noise-on-acceleration prior as presented by Anderson and Barfoot \cite{anderson_iros15}, which is defined by the following kinematic equations:
\begin{equation} \label{eq:se3_prior}
\begin{split}
\dot{\mathbf{T}}(t)&=\mbs{\varpi}(t)^\wedge{}\mathbf{T}(t), \\
\dot{\mbs{\varpi}}&=\mathbf{w}(t),\quad \mathbf{w}(t) \sim \mathcal{GP}(\mathbf{0}, \mathbf{Q}_c\delta(t-t')),
\end{split}
\end{equation}
where \(\mbf{w}(t) \in \mathbb{R}^6\) is a zero-mean, white-noise Gaussian process, and the operator, $\wedge$, transforms an element of $\mathbb{R}^6$ into a member of Lie algebra, $\mathfrak{se}(3)$ \cite{Barfoot2017}. In the interest of space, see Wong et al. \cite{wong_ral20b} for further details on this factor.

The circle factors in Figure~\ref{fig:factor_graph}\footnote{This is a general illustration with measurement factors between frames. For implementation, we associate each frame only to the reference (see (\ref{eq:meas_factor})).}, $\phi^m$, are the measurement factors defined by the lidar keypoint measurements:
\begin{equation}
\begin{split} \label{eq:meas_factor}
  \phi^m (\mbf{z}_k^\ell | \mbf{x}_\tau, \mbf{x}_k, \mbs{\theta})& = \frac{1}{2} \left( \mbf{z}_k^\ell - \mbf{g}(\mbf{x}_\tau,\mbf{x}_k) \right)^T \\ &\times \mbf{W}_k^\ell \left( \mbf{z}_k^\ell - \mbf{g}(\mbf{x}_\tau,\mbf{x}_k) \right) - \ln\left| \mbf{W}_k^\ell \right|,
\end{split}
\end{equation}
where we use the log-likelihood of a Gaussian as the factor, and $\mbf{W}_k^\ell$ is the inverse covariance matrix corresponding to measurement $\mbf{z}_k^\ell$. The keypoint, $\mbf{z}_k^\ell$, its inverse covariance matrix, $\mbf{W}_k^\ell$, and the measurement model, $\mbf{g}(\cdot)$, are quantities that depend on the network parameters, $\mbs{\theta}$. These quantities will be further explained in Section \ref{subsec:network}. 

We apply \ac{EM} (see Section \ref{sec:esgvi}) to jointly optimize for the posterior, $q(\mbf{x})$, and the network parameters, $\mbs{\theta}$, under the single objective in (\ref{eq:functional}). We emphasize that we do not use any form of groundtruth, such as pose estimates from a global positioning system, for training $\mbs{\theta}$. We learn only from the on-board lidar data.

However, \ac{ESGVI} parameter learning is a general framework based on factor graph optimization and can accommodate additional data sources beyond those presented in this work. For example, supervision from a global positioning system could be applied as unary factors for the poses of the posterior. Alternatively, a weaker form of supervision can come from applying a known measurement factor for inertial measurement unit (IMU) data, which could be used at both train and test time for a lidar-IMU odometry solution.

\subsection{Network} \label{subsec:network}
We adapt the network architecture of Barnes and Posner \cite{barnes2020} for pointclouds. They present a U-Net \cite{unet} style convolutional encoder-multi-decoder network architecture that outputs keypoints and descriptors from radar data projected into a \ac{2D} bird's-eye view image, and thus the convolutional kernels have a spatial extent in \ac{2D} Euclidean space. We achieve an equivalent effect for 3D pointclouds with KPConv \cite{thomas2019KPConv}, a pointcloud convolution method that uses kernel points arranged in a sphere of fixed radius\footnote{Thomas et al. \cite{thomas2019KPConv} also present a deformable kernel implementation, but we do not apply it in our work.}. Figure \ref{fig:network} shows the network architecture, where in place of \emph{pixels} of an image with feature channels, we have \emph{points} of a pointcloud with feature channels.

The input to the network is a lidar frame pointcloud with a channel for intensity data. Each network layer $j$, including the input layer $0$, uniformly subsamples the pointcloud into a voxel grid of dimension $dl_j$. Successive layers in the encoder increase the grid dimension by a factor of 2, i.e., $dl_{j+1} = 2\, dl_j$, and therefore the convolutions are applied at different scales in Euclidean space. The opposite is true for the decoder layers in order to have the input and output dimensions be equal. We set $dl_0$ to be $0.3$ m.

Each layer of the encoder consists of two KPConv variations of bottleneck ResNet blocks \cite{resnet}, followed by a strided variation for spatial dimension reduction. Each layer of the decoder consists of a nearest upsample operation, for spatial dimension enlargement, and a single KPConv bottleneck ResNet block. These convolution blocks apply the leaky ReLU nonlinearity and batch normalization. We direct readers to the KPConv publication \cite{thomas2019KPConv} for detailed definitions of the various block operations. As in U-Net\cite{unet}, we use skip connections between encoder and decoder layers.

The network outputs descriptor vectors, inverse measurement covariance matrices, and detector scores for each input point. These outputs are used to compute the keypoints, $\mbf{z}_k^\ell$, their corresponding inverse covariance matrices, $\mbf{W}_k^\ell$, and the output of the measurement model, $\mbf{g}(\cdot)$ (see Section \ref{subsec:problem}).

The descriptor vectors are computed for each point in the input layer and are composed of the output feature channels of all but the last encoder layer. The channel output dimension of the first layer is 64, and doubles for each subsequent layer. The output channels of the layers are concatenated with nearest upsampling to create descriptors of length 960, which are normalized into unit vectors.

The inverse measurement covariances are derived from the output of one of the two decoders (top decoder of Figure \ref{fig:network}). Applying a $1 \times 1$ linear convolution to the last decoder layer gives a final output with 6 channels. We compose the output values into inverse covariance matrices, $\mbf{W} \in \mathbb{R}^{3\times3}$, with the approach of Liu et al. \cite{liu2018deep}, which uses the following LDU decomposition for symmetric, positive definite matrices:
\begin{equation} \label{eq:LDL}
\mbf{W} = \begin{medsize}
  \bbm 1 & 0 & 0 \\ \ell_1 & 1 & 0 \\ \ell_2 & \ell_3 & 1 \ebm
  \bbm \exp d_1 & 0 & 0 \\ 0 & \exp d_2 & 0 \\ 0 & 0 & \exp d_3 \ebm
  \bbm 1 & 0 & 0 \\ \ell_1 & 1 & 0 \\ \ell_2 & \ell_3 & 1 \ebm^T, 
\end{medsize} 
\end{equation}
where $\left[ \ell_1, \ell_2, \ell_3, d_1, d_2, d_3 \right]$ is the 6D output for each point.

The detector scores are the output of the remaining decoder. After applying a $1 \times 1$ linear convolution to the last layer, the final output has 1 channel, i.e. a scalar detector score for each point. The pointcloud is then partitioned into voxels of grid size $dg$ (we set $dg$ to be $1.6$ m) for the purpose of computing one keypoint per voxel. For each voxel, we apply a softmax function over the detector scores, resulting in weights we use to compute the keypoint's coordinates along with its descriptor and inverse covariance. For example, the $\ell$th keypoint coordinate in frame $k$ is
\begin{equation} \label{eq:keypoint}
  \mbf{z}_k^\ell = \sum_{i=1}^M \frac{\exp s_i}{\sum_{j=1}^M \exp s_j} \mbf{p}^i,
\end{equation}
where $s_1, \dots, s_M$ are the detector scores of voxel $\ell$, and $\mbf{p}^1, \dots, \mbf{p}^M \in \mathbb{R}^3$ are the corresponding point coordinates. A similar computation is done to get the  descriptor vector, $\mbf{d}_k^\ell$, and inverse covariance, $\mbf{W}_k^\ell$, for each keypoint. For the inverse covariance, we apply the weighted summation over the 6D vector, and compose it into the $3 \times 3$ matrix afterward.

We use the keypoint descriptor for data association, which will be matched to a point in the reference pointcloud at time $t_\tau$. Differentiability is maintained by approximating all matches with a softmax \cite{barnes2020,Wang2019DCP}. We compute the dot product between each keypoint descriptor and all descriptors of the reference pointcloud:
\begin{equation}
  \mbf{c}_k^{\ell^T} = \mbf{d}^{\ell \, T}_k \bbm \mbf{d}_{\tau}^1 & \dots & \mbf{d}_{\tau}^N \ebm,
\end{equation}
where $\mbf{d}^\ell_k$ is the descriptor vector of keypoint $\mbf{z}^\ell_k$, and $\mbf{d}_{\tau}^1, \dots, \mbf{d}_{\tau}^N$ are the descriptor vectors of the $N$ points in the reference frame. We apply a softmax function on $\mbf{c}_k^\ell$, and compute a weighted summation. The reference point match for keypoint $\mbf{z}^\ell_k$ is therefore
\begin{equation}
  \mbf{r}_{\tau}^{\ell_k} = \sum_{i=1}^N \frac{\exp c_{k,i}^\ell}{\sum_{j=1}^N \exp c_{k,j}^\ell} \mbf{p}^i_{\tau},
\end{equation}
where $c_{k,1}^\ell, \dots, c_{k,N}^\ell$ are the scalar elements of $\mbf{c}_k^\ell$, and $\mbf{p}^1_\tau, \dots, \mbf{p}^N_\tau$ are the reference point coordinates.

We can now fully define the measurement factor in (\ref{eq:meas_factor}) with outputs of the network:
\begin{equation}
\begin{split}
  \phi^m&(\mbf{z}_k^\ell | \mbf{x}_\tau, \mbf{x}_k, \mbs{\theta}) = \frac{1}{2} \left( \mbf{z}_k^\ell - \mbf{D}\mbf{T}_{k,0} \mbf{T}_{0,\tau} \bbm \mbf{r}_{\tau}^{\ell_k} \\ 1 \ebm \right)^T \\ &\times \mbf{W}_k^\ell \left( \mbf{z}_k^\ell - \mbf{D}\mbf{T}_{k,0} \mbf{T}_{0,\tau} \bbm \mbf{r}_{\tau}^{\ell_k} \\ 1 \ebm \right) - \ln\left| \mbf{W}_k^\ell \right|,
\end{split}
\end{equation}
where $\mbf{D}$
is a $3 \times 4$ constant projection matrix that removes the homogeneous element.

Figure \ref{fig:analysis} shows visualizations of the learned detector scores and covariances. The detector favours points on, and in the vicinity of, structure such as wall corners and vertical posts. Interestingly, the nearby ground is favoured over vehicles, possibly due to their dynamic nature. We visualize sphericity \cite{thomas2018semantic} (see Section \ref{subsec:outlier}) to demonstrate the covariance. Instead of manually choosing the error metric (e.g., point-to-plane), the network adapts to low-level geometry.

\subsection{Training and Inference} \label{subsection:train_inf}
The general idea of training and inference in the \ac{ESGVI} parameter learning framework using \ac{EM} is presented in Section \ref{sec:esgvi}. In the E-step, we hold all network parameters, $\mbs{\theta}$, fixed and optimize for the posterior, $q(\mbf{x})$. In the M-step, we hold the posterior, $q(\mbf{x})$, fixed and optimize for the network parameters, $\mbs{\theta}$. Critically, the M-step does not have to be computed to completion (i.e., convergence), before alternating to the E-step, to satisfy the iterative update scheme of the data likelihood. When the M-step is not computed to completion, the algorithm is referred to as Generalized \ac{EM} (GEM).

We adapt GEM to seamlessly fit into conventional network training (i.e., stochastic gradient optimization) by including the E-step in the forward propagation routine. A window of sequential lidar frames is treated as a mini-batch of data, and forward propagation involves the following steps: 
\begin{enumerate}
\item Evaluate the lidar features and other associated outputs of each lidar frame (see Section \ref{subsec:network}).
\item Construct the motion prior, $\phi^p$, and measurement, $\phi^m$, factors (see Section \ref{subsec:problem} and \ref{subsec:network}).
\item The E-step: Inference for the current best posterior estimate $q(\mbf{x})$ of the mini-batch (window) of frames.
\end{enumerate}
Wong et al. \cite{wong_ral20b} demonstrate learning $\mbf{Q}_c$ of the motion prior (see Section \ref{subsec:problem}) in the M-step, but we do not apply it in our work and manually set suitable values for urban driving.

The E-step is simply a factor graph optimization problem, and can be solved efficiently. We apply \ac{MAP} estimation by optimizing the loss functional (\ref{eq:functional}) with the Gauss-Newton algorithm, which involves taking two approximations:
\begin{itemize}
  \item Approximate the Hessian with first-order derivatives.
  \item Approximate the expectation in (\ref{eq:functional}) at only the mean of the posterior $q(\mbf{x})$.
\end{itemize}
The first approximation is commonly made in practical applications, the alternative being Newton's method which requires second-order derivatives. In the \ac{ESGVI} framework, the second approximation is reasonable under the condition that the posterior is concentrated \cite{barfoot_ijrr20}, which we find to be the case with lidar data (i.e., rich data with accurate geometry).  In future work we may revisit these approximations and return to the full \ac{ESGVI} optimizer for the E-step.

We compute backpropagation for the network parameters, $\mbs{\theta}$, on the loss functional (\ref{eq:functional}), where only the measurement factors, $\phi^m$, are affected since the motion prior factors are constant with respect to $\mbs{\theta}$. We use the spherical-cubature rule \cite{sarkka2013bayesian} to compute sigmapoints for the posterior, $q(\mbf{x})$, in order to approximate the expectation in (\ref{eq:functional}). We do not need to compute sigmapoints over the entire posterior, which can be expensive, but just the marginals for each factor \cite{barfoot_ijrr20}. Training continues until the loss functional (\ref{eq:functional}) converges. Once converged, inference can be computed on new sequences of lidar frames for odometry (i.e., the E-step).

\subsection{Outlier Rejection} \label{subsec:outlier}

Outlier rejection is an important component to improve robustness for any estimation algorithm, and is traditionally handled with M-estimation \cite{barfoot_ijrr20} in factor graph optimization. We apply M-estimation in the E-step by applying the Geman-McClure cost function on the measurement factors, $\phi^m$ (see Section \ref{subsec:problem}), when optimizing with Gauss-Newton. M-estimation is applied at both train and test time.

While the robust cost function is sufficient for the E-step, we cannot apply it to the measurement factor when backpropagating to learn the inverse measurement covariances, $\mbf{W}$, in the M-step. Instead, we apply a hard threshold on the squared Mahalanobis term in the measurement factor with the current best posterior estimate,
\begin{equation}
  \left( \mbf{z}_k^\ell - \mbf{g}(\mbf{x}_\tau,\mbf{x}_k) \right)^T \mbf{W}_k^\ell \left( \mbf{z}_k^\ell - \mbf{g}(\mbf{x}_\tau,\mbf{x}_k) \right) > \alpha,
\end{equation}
and do not backpropagate any factor terms that exceed the threshold, $\alpha$. This threshold, which we set to $4$, is only applied during training at the backpropagation step.

Our keypoint detector, adapted from Barnes and Posner \cite{barnes2020}, determines the best keypoint in each voxel partition of the pointcloud (see Section \ref{subsec:network}). This is suboptimal for our problem formulation, as it results in keypoints in uninteresting areas (e.g., the ground plane). We can compensate at test time by judging the quality of each keypoint, $\mbf{z}_k^\ell$, with the learned inverse measurement covariance, $\mbf{W}_k^\ell$. Computing the sphericity metric \cite{thomas2018semantic} using the eigenvalues of the measurement covariance, $\lambda_3/\lambda_1$, where $\lambda_1 \geq \lambda_2 \geq \lambda_3 \in \mathbb{R}$ are the eigenvalues\footnote{Eigenvalues of ${\mbf{W}_k^\ell}^{-1}$ are the reciprocals of the eigenvalues of $\mbf{W}_k^\ell$.} of the covariance ${\mbf{W}_k^\ell}^{-1}$, is a potential way to judge the quality of each keypoint.

However, we found the computation of the eigenvalues to be too inefficient in practice. Alternatively, we apply a metric that achieves a similar effect using the diagonal elements of ${\mbf{W}_k^\ell}$ (see (\ref{eq:LDL})). We define this metric with a threshold as
\begin{equation}
  \nicefrac{\exp d_{\mathrm{min}}}{\exp d_{\mathrm{max}}} = \exp \left( d_{\mathrm{min}} - d_{\mathrm{max}} \right) < \beta,
\end{equation}
where $d_{\mathrm{min}}$ and $d_{\mathrm{max}}$ are the smallest and largest of the diagonal elements, respectively. We do not use keypoints less than the threshold, $\beta$. We found through experimentation that this metric works well on planar surfaces that are axis-aligned to the sensor frame. This threshold, which we set to $0.05$, is only applied in the E-step, i.e., we still backpropagate keypoints less than the threshold for covariance learning.

\section{Experimental Results}
\label{sec:experiments}

\subsection{Experiment Setup}
We evaluate lidar odometry on two datasets, each with a different lidar sensor. The KITTI odometry benchmark \cite{Geiger2012} has 22 sequences of Velodyne HDL-64 data collected at $10$ Hz. The first 11 (00-10) are provided as the training set, and the remaining 11 (11-21) are provided without groundtruth and act as the benchmark. Following existing work \cite{li2019,cho2020unsupervised}, we split the first 11 sequences into training and testing sequences and evaluate against the provided groundtruth.

\begin{table}[h]
  \caption{A comparison of our odometry method to those that fully learn the estimator with a deep network. DeepLO \cite{cho2020unsupervised} and our method are trained unsupervised, while LO-Net \cite{li2019} is trained with supervision from the groundtruth trajectory. Our method and LO-Net trained on sequences 00-06, while DeepLO trained on sequences 00-08. Using the KITTI odometry benchmark metric \cite{Geiger2012}, the average translation ($\%$) and orientation ($^\circ/100$ m) errors over lengths of $100$ m to $800$ m are presented. The average over sequences 00-08 are presented, as DeepLO does not present them individually. The best results are in bold.}
  \label{tab:learned-only}

  \begin{center}
  \begin{tabular}{@{}c|ccc@{}}
  \toprule
  \multirow{2}{*}{Seq.}     & Ours & DeepLO \cite{cho2020unsupervised} & LO-Net \cite{li2019} \\
   & (Unsupervised) & (Unsupervised) & (Supervised) \\ \midrule
  00-08   & \tbf{0.82}/\tbf{0.32} & 3.68/0.87 & 1.27/0.67 \\
  09      & \tbf{0.97}/\tbf{0.34} & 4.87/1.95 & 1.37/0.58 \\
  10      & \tbf{1.38}/\tbf{0.51} & 5.02/1.83 & 1.80/0.93 \\ \midrule
  Avg.    & \tbf{0.89}/\tbf{0.34} & 3.91/1.06 & 1.33/0.69 \\
  \bottomrule
  \end{tabular}
  \end{center}
  \vspace*{-0.2in}
\end{table}

Velodyne HDL-32 sensors were introduced to the Oxford RobotCar dataset \cite{Maddern2017} in the radar dataset extension \cite{barnes2020oxford}. Two $20$ Hz HDL-32 sensors (left and right) were placed on the roof of the data collection vehicle. We opted for the simpler setup of only evaluating odometry using one (left) of the two sensors. The dataset contains $30$ sequences, each $9$ km in length, and $2$ shorter sequences. All sequences were collected from a similar driving route over a week, and thus there is little variation for lidar data. We evaluate odometry on $6$ of the $9$ km sequences, where $2$ of the $6$ are used for training.

KITTI preprocessed the lidar data to account for motion distortion. While the Oxford dataset does not motion-compensate the data, we chose to not account for this effect as it is not the focus of this work, and the faster spin-rate alleviates this problem to some degree. We demonstrated motion compensation in past work with the same estimator \cite{tang_crv18,tang_ral19}, and note that it is applicable to this work as well.

\begin{table}[h]
  \caption{A comparison of our method to the current state of the art for lidar odometry methods. Using the KITTI odometry benchmark metric \cite{Geiger2012}, the average translation ($\%$) and orientation ($^\circ/100$ m) errors over lengths of $100$ m to $800$ m are presented. The best results are in bold.}
  \label{tab:everything}

  \begin{center}
  \begin{tabular}{@{}c|ccc@{}}
  \toprule
  \multirow{2}{*}{Seq.}    & Ours & LO-Net+Mapping (\ac{ICP})\cite{li2019} & LOAM \cite{zhang2017low} \\
   & (Unsupervised) & (Supervised) & (Non-Learned) \\ \midrule
  00$^\dagger$   & 0.92/\tbf{0.39} & \tbf{0.78}/0.42 & \tbf{0.78}/- \\
  01$^\dagger$   & \tbf{1.30}/\tbf{0.28} & 1.42/0.40 & 1.43/- \\
  02$^\dagger$   & 1.11/\tbf{0.42} & 1.01/0.45 & \tbf{0.92}/- \\
  03$^\dagger$   & 0.77/\tbf{0.38} & \tbf{0.73}/0.59 & 0.86/- \\
  04$^\dagger$   & 0.62/\tbf{0.22} & \tbf{0.56}/0.54 & 0.71/- \\
  05$^\dagger$   & 0.68/\tbf{0.30} & 0.62/0.35 & \tbf{0.57}/- \\
  06$^\dagger$   & \tbf{0.50}/\tbf{0.17} & 0.55/0.33 & 0.65/- \\ \midrule
  07$^*$   & \tbf{0.49}/\tbf{0.33} & 0.56/0.45 & 0.63/- \\ 
  08$^*$   & \tbf{1.01}/\tbf{0.36} & 1.08/0.43 & 1.12/- \\
  09$^*$   & 0.97/\tbf{0.34} & \tbf{0.77}/0.38 & \tbf{0.77}/- \\
  10$^*$   & 1.38/0.51 & 0.92/\tbf{0.41} & \tbf{0.79}/- \\ \midrule
  Avg. & 0.89/\tbf{0.34} & \tbf{0.82}/0.43 & 0.84/- \\
  \bottomrule 
  \end{tabular}
  \end{center}
  $^\dagger$: Sequences that our method and LO-Net train on. \\
  $^*$: Sequences that are not used for training.
\end{table}

We follow the KITTI odometry evaluation metric for all datasets, which averages the relative position and orientation errors over trajectory segments of $100$ m to $800$ m. We implemented the network using a KPConv implementation in PyTorch\footnote{https://github.com/HuguesTHOMAS/KPConv-PyTorch}. Network parameters were trained with the Adam optimizer \cite{Kingma2014}, and always trained from random initialization. Pointclouds were augmented during training with random rotations in the $z$-axis for more variation. The estimator (Gauss-Newton) was implemented using STEAM\footnote{https://github.com/utiasASRL/steam}, a C++ optimization library. Loop closures were not implemented. 

Our current implementation\footnote{On an Nvidia Tesla V100 GPU and 2.2 GHz Intel Xeon CPU.} is not real-time for a HDL-64, taking on average $359$ ms for a window of 4 frames. KPConv is a bottleneck, taking $180$ ms for pre-processing and $43$ ms for forward propagation for each frame\footnote{Only computed for the latest frame since previous ones are saved.}. Updates are in the works to improve runtime. Data association for each frame takes $19$ ms ($\times 3$ for window of 4), while Gauss-Newton takes $58$ ms. Gauss-Newton is the only C++ implementation and runs on the CPU. The rest is overhead.

\subsection{Odometry results}
We train and evaluate odometry with a window of 4 frames for KITTI. The relative pose between the latest two frames of the window are taken as the odometry output, reflecting online operation (i.e., does not use data from future frames). We compare to the current state-of-the-art methods that fully learn the estimator: LO-Net \cite{li2019} and DeepLO \cite{cho2020unsupervised}. Since DeepLO only presents the average of sequences 00-08, Table \ref{tab:learned-only} presents the results in the same way. DeepLO train on sequences 00-08, while we follow LO-Net and train on sequences 00-06. DeepLO does not perform as well as LO-Net, but has the advantage that it is unsupervised. Our method maintains the advantage of being unsupervised, and achieves better performance than both methods.

\begin{figure}[h]
  \includegraphics[width=0.45\textwidth]{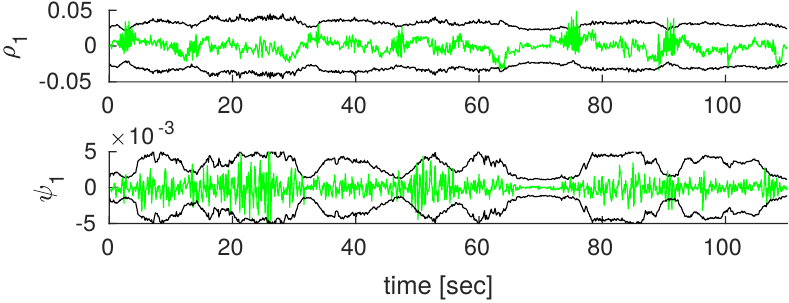}
  \caption{Odometry error of KITTI sequence $07$ with $3\sigma$ variance envelopes. In the interest of space, only two dimensions, $\rho_1$ and $\psi_1$, are shown (See (\ref{eq:pose_error})). Our estimator is in general consistent, but at times slightly overconfident.}
  \label{fig:error_plot}
  \vspace*{-0.1in}
\end{figure}

The uncertainty output of our estimator is in Figure \ref{fig:error_plot}, which shows the relative pose error of sequence $07$ with $3\sigma$ variance envelopes. Errors are computed as
\begin{equation} \label{eq:pose_error}
\mbs{\xi}_{k,k-1} = \left [ \rho_1 \: \rho_2 \: \rho_3 \: \psi_1 \: \psi_2 \: \psi_3 \right]^T = \ln \left( \mbf{T}_{k,k-1} \mbf{T}_{k,k-1}^{\mathrm{gt}^{-1}} \right)^\vee,
\end{equation}
where $\mbf{T}_{k,k-1}$ is the relative pose estimate between frames $t_k$ and $t_{k-1}$, $\mbf{T}_{k,k-1}^{\mathrm{gt}}$ is the groundtruth, $\ln(\cdot)$ is the inverse exponential map, and $\vee$ is the inverse of the $\wedge$ operator \cite{Barfoot2017}.

Table \ref{tab:everything} compares our odometry to the current state of the art for lidar odometry, which are \ac{ICP}-based methods. \ac{LOAM}\footnote{The orientation results for \ac{LOAM} are not provided in their publication.} \cite{zhang2017low} is currently leading the KITTI benchmark leaderboard, and LO-Net+Mapping is the \ac{ICP} solution presented by Li et al. \cite{li2019} that applies point masks trained with LO-Net\footnote{Our understanding is that the masks are trained without supervision from mask targets, but with supervision from groundtruth trajectory \cite{li2019}.} and manually computed surface normals. Overall, we demonstrate that our method is comparable to the current state of the art. Compared to LO-Net+Mapping, our method is learned unsupervised and does not rely on \ac{ICP} for data association. Compared to \ac{LOAM}, our method can easily be tuned for different platforms and lidar sensors by learning from just the on-board lidar data.

Our submission to the online benchmark, with the same network and parameters, achieved $1.07 \%$ average translation and $0.36^\circ/100$ m average orientation error. In comparison, \ac{LOAM} currently has $0.55 \%$ translation and $0.13^\circ/100$ m orientation ($0.88 \%$ translation in original publication \cite{zhang2017low}). Our results are more comparable to \ac{SuMa} \cite{behley2018rss} ($1.39 \%$ and $0.34^\circ/100$ m) and \ac{SuMa}++ \cite{chen2019iros} ($1.06 \%$ and $0.34^\circ/100$ m), which are both well-regarded \ac{ICP}-based methods. DeepLO and LO-Net currently do not have submissions. Considering we do not apply loop closure, which the benchmark permits, we believe our method achieved reasonable performance.

\begin{figure}[]
  \includegraphics[width=0.45\textwidth]{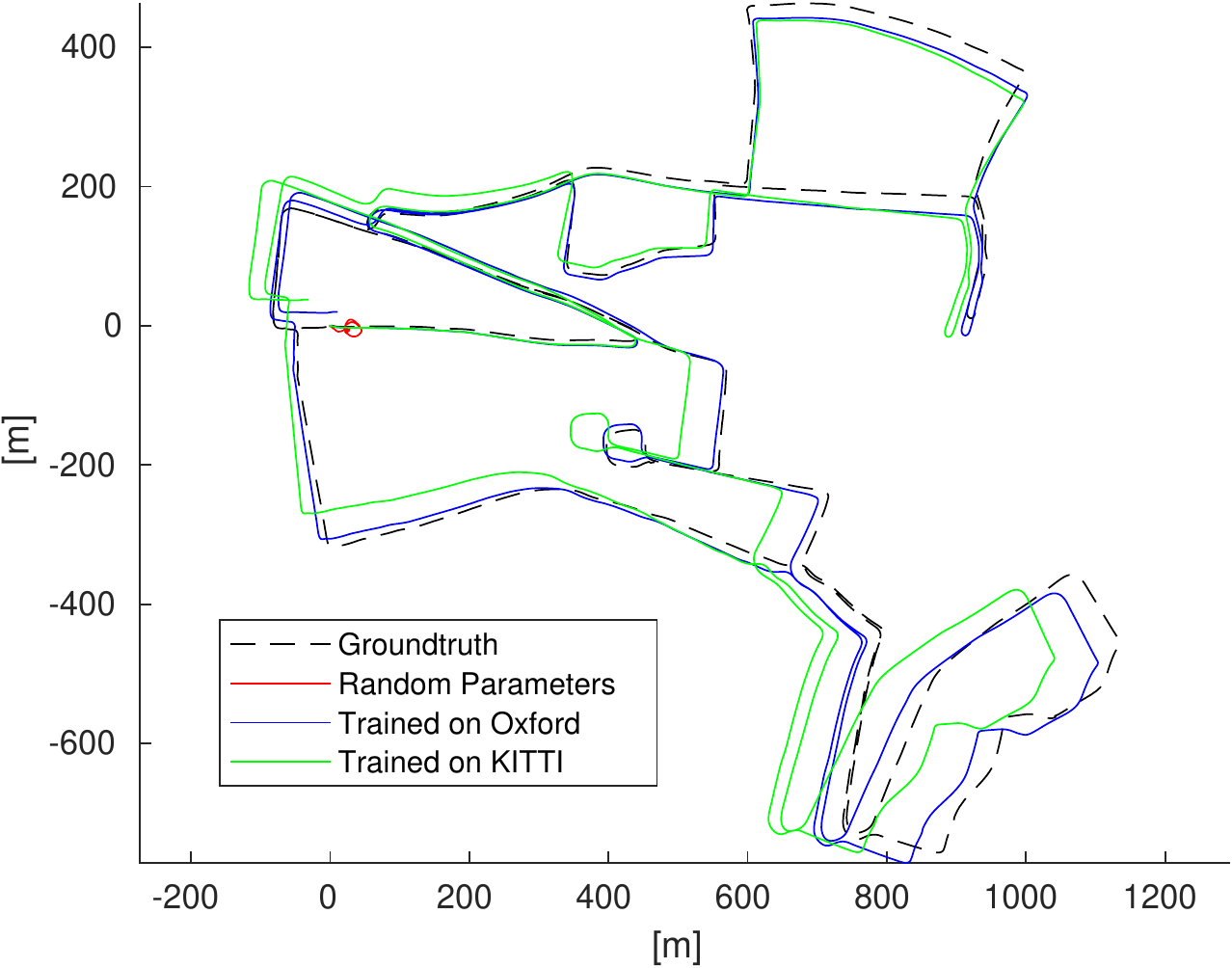}
  \caption{Odometry paths for sequence 2019-01-15-14-24-38 of the Oxford RobotCar dataset \cite{Maddern2017,barnes2020oxford} with a Velodyne HDL-32. We compare against the performance when using a network trained on a different dataset, KITTI \cite{Geiger2012}, which is a Velodyne HDL-64. Odometry fails prematurely due to numerical instability when the network is not trained (red).}
  \label{fig:oxford_path}
 \vspace*{-0.2in}
\end{figure}

We demonstrate automated tuning by training and testing on the Oxford dataset, and comparing it to performance of the network trained on KITTI (i.e., the network applied in Tables \ref{tab:learned-only} and \ref{tab:everything}). We optimize over a window of 7 and use the lidar data at $10$ Hz (i.e., skip every other frame), settings with which the KITTI trained network performed best. The same settings were applied when training on the Oxford dataset. The results in Table \ref{tab:oxford_table} show a clear improvement when parameters are trained with data related to the deployment. Figure \ref{fig:oxford_path} shows a qualitative plot of the odometry paths for sequence 2019-01-15-14-24-38, where we additionally show the performance with an untrained network.

\begin{table}[h]
  \caption{Odometry results for our method on Velodyne HDL-32 data of the Oxford RobotCar dataset \cite{Maddern2017,barnes2020oxford}. Using the KITTI odometry benchmark metric \cite{Geiger2012}, the average translation ($\%$) and orientation ($^\circ/100$ m) errors over lengths of $100$ m to $800$ m are presented. The best results are in bold.}
  \label{tab:oxford_table}

  \begin{center}
  \begin{tabular}{@{}c|ccc@{}}
  \toprule
  \multirow{1}{*}{Seq.}  & Trained on Oxford & Trained on KITTI       \\ \midrule
  2019-01-10-11-46-21$^\dagger$   & \tbf{2.85}/\tbf{1.29} & 3.21/1.55 \\
  2019-01-18-15-20-12$^\dagger$   & \tbf{2.41}/\tbf{1.13} & 3.04/1.47 \\ \midrule
  2019-01-15-13-06-37$^*$         & \tbf{2.48}/\tbf{1.20} & 2.89/1.43 \\
  2019-01-15-14-24-38$^*$         & \tbf{2.60}/\tbf{1.25} & 2.95/1.51 \\
  2019-01-16-13-09-37$^*$         & \tbf{2.56}/\tbf{1.20} & 3.15/1.48 \\
  2019-01-16-14-15-33$^*$         & \tbf{2.99}/\tbf{1.47} & 3.60/1.76 \\ \midrule
  Avg.                            & \tbf{2.65}/\tbf{1.26} & 3.14/1.53 \\
  \bottomrule
  \end{tabular}
  \end{center}
  $^\dagger$: Sequences that we train on (applicable only to 'Trained on Oxford'). \\
  $^*$: Sequences that are not used for training.
  \vspace*{-0.2in}
\end{table}

\subsection{Ablation Study}
Table \ref{tab:ablation} shows the results of an ablation study, where we remove various components of our method. In addition to the KITTI benchmark metrics for translation and orientation, we compute an average Mahalanobis distance metric \cite{brossard2020new},
\begin{equation} \label{eq:avg_mah}
  \left( \sum_{k=1}^{K} \frac{\mbs{\xi}_{k,k-1}^T \mbf{Q}_{k,k-1}^{-1}\mbs{\xi}_{k,k-1}}{\mathrm{dim}(\mbs{\xi}_{k,k-1})K} \right)^{1/2},
\end{equation}
where $\mbs{\xi}_{k,k-1}$ is the error as defined in (\ref{eq:pose_error}) and $\mbf{Q}_{k,k-1}$ is the corresponding covariance. A value close to $1$ is ideal.

`No Sampling' refers to evaluating the expectation in the loss functional (\ref{eq:functional}) at only the mean of the posterior in the M-step (see Section \ref{subsection:train_inf}). We see that approximating the expectation with sigmapoints is insignificant for this problem, which is consistent with the approximation made in the E-step. `No $\beta$' and `No $\alpha$' refers to not applying the $\beta$ and $\alpha$ thresholds in Section \ref{subsec:outlier}, which clearly perform worse for translation and orientation. The exception is the Mahalanobis metric for `No $\alpha$', which performs the most consistently and more conservatively than the rest. Backpropagating outliers means the learned uncertainties must account for them, thus it makes sense for the estimator to become more conservative.

\begin{table}[h]
  \caption{An ablation study over components of our method on the KITTI odometry dataset. Using the KITTI odometry benchmark metric \cite{Geiger2012}, the average translation ($\%$) and orientation ($^\circ/100$ m) errors over lengths of $100$ m to $800$ m are presented. We additionally compute the average squared Mahalanobis distance (third metric in each column) of the relative pose estimates (see (\ref{eq:avg_mah})), which ideally is $1$ for a consistent estimator.}
  \label{tab:ablation}
  \centering
  \resizebox{0.49\textwidth}{!}{%
  \begin{tabular}{@{}c|cccc@{}}
  \toprule
  \multirow{1}{*}{Seq.}   & Full method & No Sampling & No $\beta$ & No $\alpha$ \\ \midrule
  07   & 0.49/0.33/1.22 & \tbf{0.48}/\tbf{0.29}/1.23 & 0.61/0.38/1.67 & 1.22/0.96/\tbf{1.06} \\
  08   & 1.01/0.36/2.66 & \tbf{0.96}/\tbf{0.33}/2.71 & 1.17/0.45/6.98 & 2.23/0.84/\tbf{2.09} \\
  09   & \tbf{0.97}/\tbf{0.34}/1.31 & 0.98/0.36/1.34 & 1.36/0.56/1.98 & 2.44/0.92/\tbf{1.14} \\
  10   & \tbf{1.38}/\tbf{0.51}/1.54 & 1.56/0.58/1.55 & 2.13/0.84/2.09 & 2.13/1.58/\tbf{1.28} \\ \midrule
  Avg. & \tbf{0.96}/\tbf{0.38}/1.68 & 0.99/0.39/1.71 & 1.32/0.56/3.18 & 2.00/1.07/\tbf{1.39} \\
  \bottomrule
  \end{tabular}
  }
  \vspace*{-0.2in}
\end{table}

\section{Conclusion and Future Work}
\label{sec:conclusion}
In this paper, we presented the first application of deep network parameter learning for \ac{ESGVI}. We showed that our parameter learning framework can learn the parameters of a deep network without groundtruth supervision. Our application to lidar odometry resulted in performance comparable to current state-of-the-art \ac{ICP}-based methods, while being simple to train for new deployments with different lidars.

For future work, we plan on extending parameter learning for \ac{ESGVI} to other rich sensors (i.e., camera, radar). We are interested in estimation problems beyond odometry, and will focus on localization and mapping. Unfortunately, the current odometry implementation is incapable of running in real-time. The current computational bottleneck is the pointcloud convolution operator, KPConv \cite{thomas2019KPConv}. A real-time implementation of KPConv is in the works such that we believe real-time performance is easily achievable.\vspace*{-0.05in}






\bibliographystyle{bib/IEEEtran}
\bibliography{bib/refs}

\end{document}

%% file: notation_header.tex
\newcommand{\bbm}{\begin{bmatrix}}
\newcommand{\ebm}{\end{bmatrix}}
\newcommand{\mbf}{\mathbf}
\newcommand{\tbf}{\textbf}
\newcommand{\mbs}[1]{{\boldsymbol{#1}}}

\newcommand{\beq}{\begin{equation}}
\newcommand{\eeq}{\end{equation}}
\newcommand{\bdis}{\begin{displaymath}}
\newcommand{\edis}{\end{displaymath}}
\newcommand{\beqn}[1]{\begin{subequations}\label{eq:#1}\begin{eqnarray}}
\newcommand{\eeqn}{\end{eqnarray}\end{subequations}}